\documentclass[10pt,twocolumn,letterpaper]{article}

\usepackage{wacv}              

\definecolor{wacvblue}{rgb}{0.21,0.49,0.74}
\usepackage[pagebackref,breaklinks,colorlinks,allcolors=wacvblue]{hyperref}
\usepackage{graphicx}
\usepackage{amsmath}
\usepackage{amssymb}
\usepackage{booktabs}
\usepackage{tabularx}
\usepackage{multirow}
\usepackage{float}
\usepackage{caption}
\usepackage[flushleft]{threeparttable}

\usepackage{paralist}

\usepackage{amssymb} 
\usepackage{pifont}  
\usepackage{colortbl}
\usepackage{color}
\usepackage{xcolor}
\usepackage{url}
\newcommand{\cmark}{{\color{black}\ding{51}}}
\newcommand{\xmark}{{\color{gray}\ding{55}}}
\usepackage[pagebackref,breaklinks,colorlinks]{hyperref}
\usepackage{subcaption}

\usepackage[capitalize]{cleveref}
\crefname{section}{Sec.}{Secs.}
\Crefname{section}{Section}{Sections}
\Crefname{table}{Table}{Tables}
\crefname{table}{Tab.}{Tabs.}
\crefname{figure}{Fig.}{Figs.}

\def\wacvPaperID{1449} 
\def\confName{WACV}
\def\confYear{2026}

\title{LENVIZ: A High-Resolution Low-Exposure Night Vision Benchmark Dataset\vspace{-0.5em}}

\author{
Manjushree Aithal$^1$, Rosaura G. VidalMata$^1$, Manikandtan Kartha, Gong Chen$^1$, Eashan Adhikarla$^2$,\\ 
Lucas N. Kristen$^3$, Zhicheng Fu$^1$, Nikhil A. Madusudhana$^1$, and Joe Nasti$^1$\\ 
{\tiny ~ }\vspace{-0.5em} \\
$^1$ Lenovo Research, $^2$ Lehigh University, $^3$ Motorola Mobility \\
\tt\small \{maithal, rosaurav, gochen24, zcfu, amnikhil, jnasti\}@lenovo.com, \\
\tt\small eaa418@lehigh.edu, lucask@motorola.com \\
\tt\small \url{https://github.com/rosauravidal/LENVIZ}
}

\makeatletter
\let\@oldmaketitle\@maketitle
\renewcommand{\@maketitle}{\@oldmaketitle
    \begin{center}
    \begin{minipage}{\textwidth}
        \centering
        \vspace{-1.5em} \includegraphics[width=\textwidth]{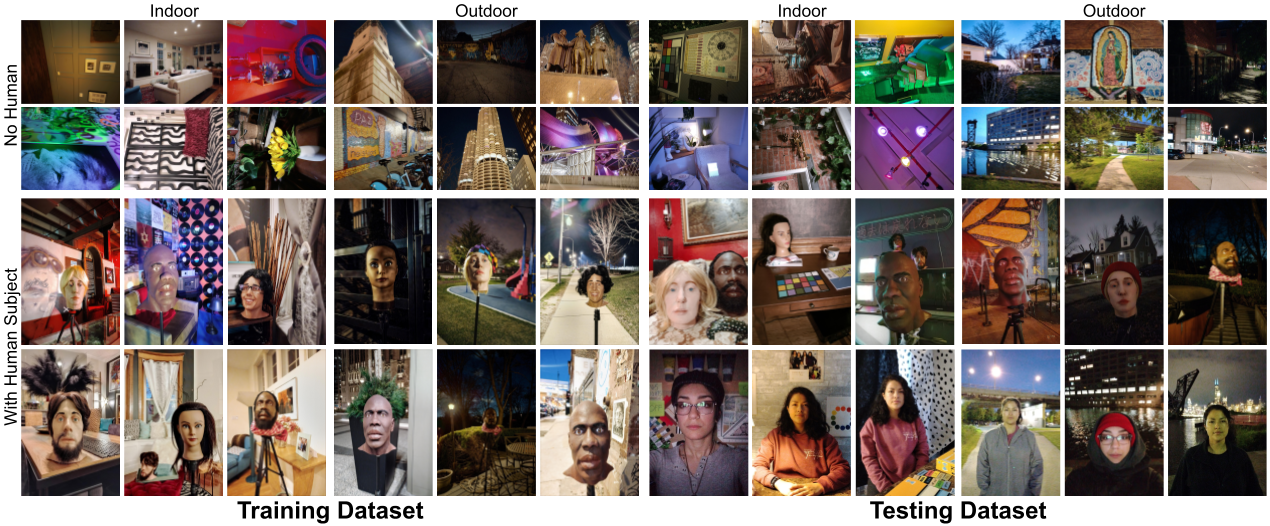}\vspace{-2mm}
        \captionof{figure}{Example of Low-light images in the LENVIZ dataset Training and Test partitions
        }
        \label{fig:teaser}\vspace{2mm}
    \end{minipage}
    \end{center}
}
\makeatother

\begin{document}
\maketitle 
\begin{abstract}
Low-light image enhancement is crucial for a myriad of applications, from night vision and surveillance, to autonomous driving. However, due to the inherent limitations that come in hand with capturing images in low-illumination environments, the task of enhancing such scenes still presents a formidable challenge. To advance research in this field, we introduce our \textbf{L}ow \textbf{E}xposure \textbf{N}ight \textbf{Vi}sion (\textbf{LENVIZ}) Dataset, a comprehensive multi-exposure benchmark dataset for low-light image enhancement comprising of over 230K frames showcasing 24K real-world indoor and outdoor, with-and-without human, scenes. Captured using 3 different camera modules, LENVIZ offers a wide range of lighting conditions, noise levels, and scene complexities, making it the largest publicly available up-to 4K resolution benchmark in the field. LENVIZ includes high quality human-generated ground truth, for which each multi-exposure low-light scene has been meticulously curated and edited by expert photographers to ensure optimal image quality. Furthermore, we also conduct a comprehensive analysis of current state-of-the-art low-light image enhancement techniques on our dataset and highlight potential areas of improvement. 
\end{abstract}
    
\vspace{-2em}\section{Introduction} \vspace{-0.5em}
\label{sec:intro}

Low-light imaging, critical for applications such as night vision and surveillance, presents significant challenges for traditional imaging systems. Due to their limited dynamic range, increased noise, reduced signal-to-noise ratio, and potential image artifacts, these systems struggle to capture high quality images in low-illumination environments. Effective low-light image enhancement techniques aim to improve the visual quality of captured scenes by increasing brightness, reducing noise, correcting exposure imbalances, and enhancing visual detail. This process often involves manipulating pixel values to restore lost information and create a more visually appealing and informative image. 

Deep learning has emerged as a powerful tool for this task, and while Convolutional Neural Networks (CNNs) 
can learn complex relationships between low-light and well-lit images, training these models requires large datasets of high-quality image pairs which can be challenging and costly to obtain. 
Early works~\cite{MIT5K, LLIE, VE-LOL, UHDLOL2023} used synthetic data with unrealistic lighting conditions, leading to models that struggle with real-world low-light scenarios. Recent efforts~\cite{LOL, SID, ExDarkDataset} focus on creating datasets of real low-light and well-lit image pairs captured under diverse lighting conditions, allowing models to learn generalizable enhancement strategies, but are still limited in size and capture conditions.  Having a comprehensive benchmark dataset is essential for advancing research in low-light image enhancement. Such a dataset should provide a diverse range of low-light scenes, capturing various lighting conditions, noise levels, and scene complexities. Moreover, high-quality ground truth images are essential for training and evaluating the performance of different enhancement methods.  Unfortunately, existing datasets~\cite{MIT5K, LOL, SID, ExDarkDataset, UHDLOL2023} often suffer from limitations in terms of size, range of capture environments, depicted subjects, or ground-truth availability.  

To address these challenges, we introduce the \textbf{Low Exposure Night Vision (LENVIZ) Dataset}, a large-scale multi-exposure benchmark for low-light image enhancement.  LENVIZ is the largest low-light benchmark datasdet to-date (234K frames), offering a comprehensive range of scene types, 3 different camera modules, and various ISP settings, addressing a critical gap in existing benchmarks. Additionally, LENVIZ provides 9 distinct exposure frames for each scene along with a long-exposure shot, enabling researchers to explore different enhancement techniques and training schemes. To ensure accurate evaluation \& training, we include photography expert-edited ground truth images. Beyond traditional variations, our dataset offers a rich diversity of elements, including different illuminance levels, capture environments, realistic mannequins for accurate skin tone representation, and facial accessories (glasses, mustaches, etc.), all of these providing more comprehensive features for training (See Fig.~\ref{fig:teaser}). 
We also introduce a curated test dataset with 1,468 frames and a total of 203 unique scenes designed by taking into account multiple benchmark platforms, including DxOMark\footnote{\url{https://corp.dxomark.com/}}. This test dataset encompasses a variety of scenes designed to evaluate not only the performance efficacy of state-of-the-art enhancement and over-exposure recovery methods but also their robustness and generalizability across diverse conditions.


LENVIZ provides a valuable resource for researchers and practitioners working on low-light image enhancement and multi-exposure fusion. By offering a comprehensive, diverse, and high-quality dataset, we enable the development and evaluation of more robust and effective techniques. To demonstrate the value of our benchmark dataset, we conduct an in-depth evaluation of state-of-the-art low-light image enhancement techniques when trained on our data. Our evaluation highlights the strengths and weaknesses of existing methods and provides insights into challenges and opportunities for future research in the field. In summary, the contributions of our paper are: 

\begin{compactitem}
    \item \textbf{Introduction of the LENVIZ Dataset:}~To the best of our knowledge, we are introducing the largest to-date (234k frame), novel low-light benchmark dataset.
    \item \textbf{Flexibility of Inputs:}~We cater 9 distinct exposure frames per scene along with a long-exposure shot, allowing for flexible input combinations.
    \item \textbf{High-Quality Ground Truth:}~Our dataset includes expert-edited ground truth images, ensuring human-based enhancement for optimal image quality. 
    \item \textbf{Standardized Exposure Calculation:}~We introduce a method to determine exposure brackets and illuminance readings using camera parameters.
    \item \textbf{Curated Test Dataset:} We constructed a comprehensive test dataset across 203 unique scenes, leveraging diverse benchmark platforms to evaluate enhancement techniques under varying conditions.
    \item \textbf{Benchmarking State-of-the-Art Techniques:}~We have conducted an in-depth evaluation of current state-of-the-art techniques in the LENVIZ dataset.
\end{compactitem} 
\begin{table*}[!ht]
\centering
\caption{Summary of existing Low-light enhancement open source datasets. 
Abbreviations: Indoor/Outdoor: Indoor (\textbf{I}), Outdoor (\textbf{O}), Frame type: Single frame (\textbf{S}), Multi-exposure frames (\textbf{M}), and Long-exposure frame (\textbf{L})}\label{tab:related_dataset} \vspace{-0.5em}
    \small
    \begin{tabular}{r|c|c|c|c|c|c}
    \hline 
    \textbf{Dataset} & \textbf{Frame Type} & \textbf{Resolution} & \textbf{Frames} & \textbf{Environment} & \textbf{Human subjects}  &  \textbf{Human GT} \\ \hline 
    MIT 5K\cite{MIT5K} 
        & S 
        & 3040x2014 
        & 5,000 
        & I,O 
        & \cmark 
        & \xmark \\ 
    
    NEF\cite{NEF}
        & S 
        & -  
        & 88
        & I,O 
        & - 
        & \xmark \\ 
    
    Phos\cite{Phos} 
        & M 
        & - 
        & 15 
        & I,O 
        & \xmark 
        & \xmark \\ 
    
    VV Dataset & 
        S & - 
        & 24 
        & I,O 
        & \xmark 
        & \xmark \\ 
    
    LLIE\cite{LLIE} 
        & M 
        & 64×64 
        & 10,000 
        & I,O 
        & \xmark 
        & \cmark \\ 
    
    SID\cite{SID} 
        & M+L 
        & \begin{tabular}[c]{@{}c@{}}4240×2832\\6000x4000\end{tabular} 
        & 424 
        & I,O 
        & \cmark 
        & \xmark \\ 

    LOL\cite{LOL} 
        & M 
        & 400x600 
        & 500 
        & I,O 
        & \xmark 
        & \xmark \\ 

    SICE\cite{SICE} 
        & M 
        & \begin{tabular}[c]{@{}c@{}}3000x2000\\6000x4000\end{tabular} 
        & 4,431 
        & I,O 
        & \cmark 
        & \xmark \\ 

    ExDark\cite{ExDarkDataset} 
        & S 
        & - 
        & 7,363 
        & I,O 
        & \cmark 
        & \cmark \\ 

    VE-LOL\cite{VE-LOL} 
        & M $\|$ \hspace{0.2em}S
        & 1080x720 
        & 2,500 $\|$ \hspace{0.1em}10,940
        & I,O 
        & \cmark 
        & \cmark \\ 

    DarkZurisch\cite{DarkZurisch} 
        & S 
        & - 
        & 5,381 
        & O 
        & \xmark 
        & \cmark \\ 

    ELD\cite{ELD} 
        & M 
        & - 
        & 60 
        & I 
        & \xmark 
        & \xmark \\ 
    
    TMDIED\cite{TMDIED2021} 
        & M 
        & - 
        & 222 
        & I 
        & \xmark 
        & \xmark \\ 
    
    RELLISUR\cite{rellisur2023} 
        & M 
        & 0.39-6.25 Mpx 
        & 2250 
        & I,O 
        & \xmark 
        & \cmark \\ 
    
    LDR\cite{LDR} 
        & M 
        & - 
        & 1800 
        & I,O 
        & \xmark 
        & \xmark \\ 
    
    UHD-LOL\cite{UHDLOL2023} 
        & M 
        & \begin{tabular}[c]{@{}c@{}}3840x2160\\7680x4320\end{tabular} 
        & \begin{tabular}[c]{@{}c@{}}8,099\\2,966\end{tabular} 
        & I,O 
        & \cmark 
        & \cmark \\ 
    
    LSRW\cite{LSRW2023} 
        & M 
        & -
        & 5,650 
        & I,O 
        & \xmark 
        & \xmark \\ 
    
    Few-shots\cite{NikonfewShot} 
        & M+L 
        & 512x512 
        & 280 
        & I,O 
        & \xmark 
        & \xmark \\ 
    
    LOM\cite{LOM2023} 
        & M 
        & 3000x4000 
        & 25 - 65 
        & I 
        & \xmark 
        & \xmark \\  
     SDE\cite{eventImage} 
        & M 
        & 346x260
        & 30,000 
        & I,O 
        & \xmark 
        & \xmark \\ 
    RLED\cite{NERNet} 
        & M 
        & -
        & 80,400
        & I,O 
        & \xmark 
        & \xmark \\ \hline 
    
    \rowcolor{gray!20}\textbf{LENVIZ}
        & \textbf{S+M+L}
        & \begin{tabular}[c]{@{}c@{}}\textbf{4080x3072}\\\textbf{3264x2448}\end{tabular}
        & \begin{tabular}[c]{@{}c@{}}\textbf{80,642}\\\textbf{ 154,046}\end{tabular}
        & \textbf{I,O}
        & \textbf{\cmark} 
        & \textbf{\cmark} \\ \hline
    \end{tabular}\vspace{-0.5em} 
\end{table*} \label{dataset_summary}

\vspace{-0.5em}\section{Related Work}\vspace{-0.5em}
The increasing demand for AI-driven solutions to address complex challenges has spurred the development of diverse, real-world datasets. This is particularly evident in the domains of low-light image enhancement and multi-exposure recovery. MIT-Adobe FiveK \cite{MIT5K} captured 5,000 unpaired RAW images under daylight and low-light conditions in indoor and outdoor settings. \cite{NEF} presented a dataset comprising 88 unpaired low-light images, of which 46 were captured using a Canon digital camera and the remainder were sourced from online websites such as Google, NASA, etc. Although this dataset exhibited a notable emphasis on local low-contrast regions with higher global illumination variation, its unpaired nature and limited size hindered its applicability for deep learning training methodologies. The Phos\cite{Phos} dataset 
provides 15 distinct scenes with 9 images captured at uniform illumination levels and 6 images under nonuniform conditions. Despite offering diverse and extensive scenes and multi-exposure intermediate captures, the limited number of images within this paired dataset restricted its utility to testing purposes.  Similar-sized datasets include
the VV-dataset\footnote{\url{https://sites.google.com/site/vonikakis/datasets}} with 24 single-exposure unpaired images, and MEF \cite{MEF} with 17 high-quality multi-exposure images, encompassing underexposed, overexposed, and intermediate illumination levels, and LIME \cite{LIME}  with low-resolution 10 unpaired low-light data. 

LLIE~\cite{LLIE} introduced a synthetically generated dataset with more than 20,000 images derived from the UCID \cite{ucid}, BSD \cite{BSD}, and Google Image Search datasets. This included 10,000 paired images, each consisting of one high-quality and one low-light natural image. As the first substantial paired low-light benchmark dataset, it offered a valuable resource for model training; however, a notable limitation was the absence of camera module specifications. The See-In-the-Dark dataset~\cite{SID}, features RAW 12 burst images captured under low-illumination conditions and paired with corresponding long-exposure high-quality images. A total of 424 unique images were acquired using Sony $\alpha$ 7s II and Fujifilm X-T2 cameras. Although the higher diversity and low-light specificity of this dataset were advantageous, its use was restricted by the requirement for RAW input images. The LOw-Light (LOL) dataset~\cite{LOL}, provides 500 paired low-light images, and SICE \cite{SICE} then introduced 4,431 multi-exposure paired images using multiple camera modules (6 different DSLR devices $+$ iPhone 6s). 

\vspace{-0.5em}This marked the start of a growing interest in low-light data which led to the introduction of several new low-light datasets. ExDark \cite{ExDarkDataset} was released, comprising 7,363 unpaired low-light images captured under twilight conditions. Although its size is substantial, its unpaired nature limits its applicability to unsupervised learning. \cite{VE-LOL} offered 10,940 unpaired images, along with 2,500 paired real-world images (1,000 synthetically generated using RAW data from RAISE dataset \cite{raise}). \cite{DarkZurisch} contributed 2,920 twilight and 2,461 night-time images captured using a GoPro Hero 5 camera. Additionally, \cite{ELD} introduced 60 paired low-light images captured with four different DSLR cameras. Another 222 paired dataset was introduced by \cite{TMDIED2021}, providing different lighting conditions (night, sunset, day, cloudy, sunlight etc). \cite{rellisur2023} introduced a dataset of 2,250 paired images, captured using a Canon EOS 6D camera equipped with a Canon 70-300mm lens. Despite the increasing diversity and realism of newly released datasets, a common limitation was the relatively small size of paired data or the prevalence of large unpaired datasets, which hindered their applicability for most model training paradigms.

\vspace{-0.5em}In 2023, a significant milestone was reached with the introduction of 4 major paired benchmark datasets offering a substantial number of low-light images. \cite{LDR} contributed 1,800 unpaired indoor and outdoor RAW images captured with varying ISOs and exposure values, focusing on low-light denoising applications. \cite{UHDLOL2023} provided 11,065 paired images, sourcing the normal lighting images from publicly available datasets~\cite{publicData} and synthetically generating the corresponding low-light images. LSRW's\cite{LSRW2023} 5,650 paired images were also released, the images captured with a Nikon D7500 camera and a HUAWEI P40 Pro phone. LSRW consisted of normal/underexposed data pairs, with a fixed ISO of 100 for normal light and 50 for low-light conditions. 
In the most recent releases, author working on event-based method such as~\cite{eventImage} provided SDE dataset that comprises of 30K paired indoor and outdoor images with dark and normal illumination levels. Additionally, ~\cite{NERNet} released some large-scale (80K), multi-illumination levels and pixel-aligned GT for low-light conditions. 

Despite these advancements, existing datasets often lack a comprehensive combination of multi-exposure capabilities, paired with long-exposure and high-quality ground-truth data, diverse scene locations, lighting ranges, scene types, scene subject types, and camera modules, particularly at up-to 4K resolution. From Table~\ref{tab:related_dataset}, we can see that while some datasets provide diverse data, they are either lacking in multi-frames inputs, dataset size, or human-generated ground-truth. 
To address these shortcomings, we analyzed the characteristics and gaps in existing low-light datasets and developed our novel LENVIZ benchmark dataset, incorporating these insights.

\vspace{-0.5em}\section{LENVIZ Capture}

Our dataset is composed of $24K$ low-light scenes captured over $3$ different camera modules. 
We aimed to create a diverse dataset suitable for pixel-level image processing tasks such as image enhancement and high dynamic range (HDR) imaging in challenging low-light conditions. With this in mind, we provide up to 9 multi-exposure frames as well as a long-exposure shot for each scene ($234,688$ total frames) to aid in the development of single, and multi-exposure image enhancement approaches.  In this section we detail the capture pipeline for our data as well as camera-specific details and configurations. Additional details on the relationship between the estimated illuminance and the exposure time are provided in Supp~\ref{fig:lux_est_image}

\subsection{Camera module Information}

To broaden our dataset's representativeness and improve the generalizability of algorithms trained using our data, we provide scenes captured over $3$ different camera modules. 
Information on the specific details (aperture, camera module-size, FOV, etc.) is provided in Supp~\ref{supp:captures}. While our S5K4H7YX03-FGX9 camera is better suited for close range captures, S5KJN1SQ03 and S5KJNS camera features allow for richer capture of details at medium and long range.



It is important to note that, the performance and characteristics of camera module can vary based on their placement on a device and intended function, making certain camera module more suitable for specific applications. 
Our first camera module, S5K4H7YX03-FGX9, was the front-facing camera on a mobile phone. These cameras are essential for selfies and video calls but are constrained in physical size due to their placement above the phone display. Although their resolution is generally lower than that of rear cameras, they are designed to capture scenes at close range (usually an arm's length) and often benefit from additional light from the screen. As such, our images from this camera module tend to contain a higher concentration of faces (distribution presentation in Supp~\ref{supp:content_class}
). The other two camera module, S5KJN1SQ03 and S5KJNS, were used as rear cameras on different devices. These are designed to serve as the primary lens for capturing the environment. Unlike front cameras, they offer higher resolution and specifications for more detailed scenes, which typically involve greater depth and lower lighting conditions. While face detection remains important for rear cameras, they are also used for capturing a wider range of subjects, such as landscapes and objects.


\begin{figure*}[ht]
   \centering
   \includegraphics[width=0.8\linewidth]{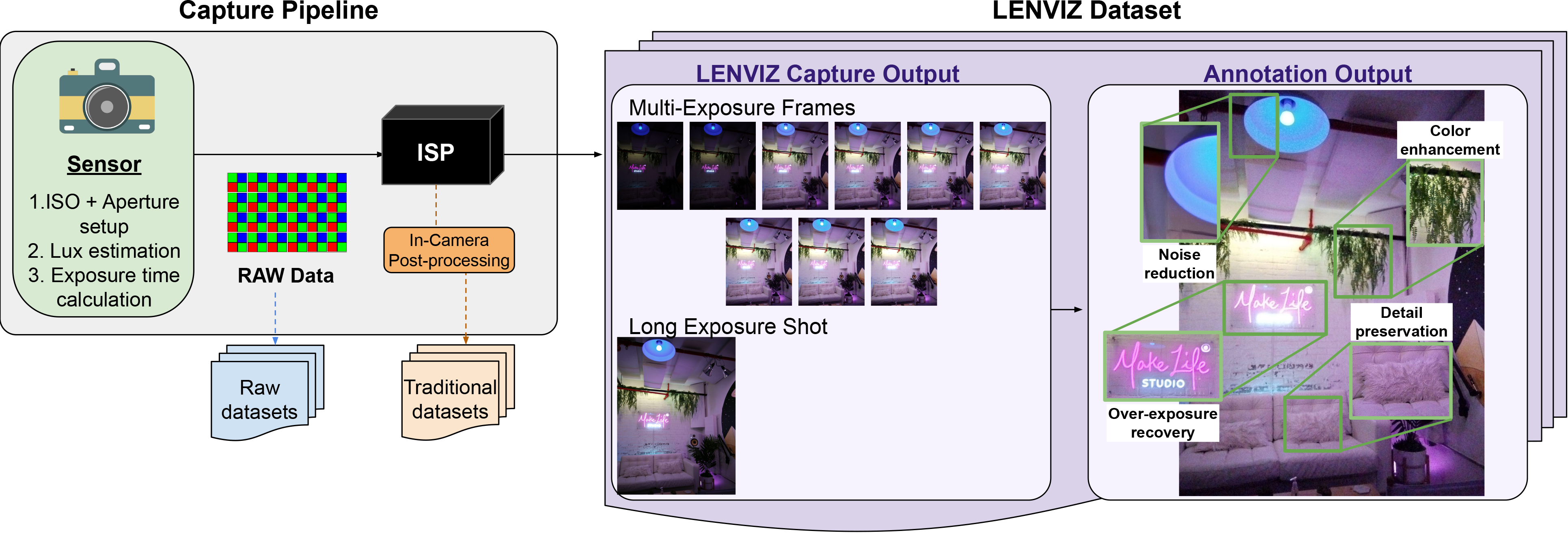}\vspace{-.75em}
    \caption{Detailed camera capture + post capture pipeline with traditional as well as LENVIZ dataset}\label{fig:capture_annotation_pipeline} \vspace{-1em}
\end{figure*}

\vspace{-0.25em}\subsection{Ambient Light Illuminance Calculation}
For our data acquisition, we implemented stringent controls over camera parameters such as ISO sensitivity, exposure time, and aperture. Concurrently, we performed an estimation of the scene illuminance levels using our capturing devices, this measurement indicates the intensity of light in a scene and is essential for characterizing low-light environments and is included for each scene. This approach ensured a highly controlled environment for image capture, thereby enhancing the dataset's suitability for benchmarking low-light image enhancement algorithms. We estimate the illuminance readings of the captured images using key camera parameters like exposure time, ISO sensitivity, and lens aperture: \vspace{-1em} 
{\small \begin{equation}
\text{L'} = \beta_{\text{illuminance}} \times f^2 \times \left( \frac{10^{11}}{t_{\text{exp}} \times ISO} \right)\label{equation 1}
\end{equation}}\vspace{-1em} 

Here, the exposure time ($t_{\text{exp}}$) determines the duration (in $ns$) that the camera module is exposed to light. $ISO$ sensitivity controls the camera module's sensitivity to light, with higher values increasing sensitivity but also introducing noise. Lens aperture ($f$) regulates the amount of light reaching the camera module (fixed for all camera modules in our dataset). Finally, camera-specific tuning constants ($\beta_{\text{illuminance}}$) account for camera module characteristics and ISP tuning. Our illuminance estimation is inversely proportional to both the exposure time and the ISO value. Longer exposure times and higher ISO values reduce the calculated illuminance. The aperture size $f$, directly influences the light-gathering capacity of the lens. Additionally, when a post-processing sensitivity boost is applied (e.g., through ISO adjustment as part of the ISP tuning of a camera module), the ISO value is modified by the post-boost percentage. 


\subsection{Long-Exposure Time Calculation}
The exposure time for our long-exposure images is computed based on the ambient light intensity, measured in illuminance, using a formula derived from the relationship between illuminance and exposure time. 
Equation~\ref{equation 2} governs the exposure time calculation for the long-exposure ground truth image given the estimated illuminance of the scene as follows:\vspace{-1em}
{\small \begin{equation}
\text{exposure\textsubscript{gt}} = \begin{cases} 
\frac{\gamma_{0}}{\text{lux}} & \text{if } \text{illuminance} < 1 \\
\frac{\gamma_{1}}{\text{lux}} & \text{if } 1 < \text{illuminance} < 8 \\
\frac{\gamma_{2}}{\text{lux}} & \text{if } 8 \leq \text{illuminance} < 10 \\
\frac{\gamma_{3}}{\text{lux}} & \text{if } 10 \leq \text{illuminance} < 15 \\
\frac{\gamma_{4}}{\text{lux}} & \text{if } \text{illuminance} \geq 15
\end{cases} \label{equation 2}
\end{equation}} \vspace{-0.5em}

The $\gamma$ values of the long-exposure settings is determined empirically and the values we used during our collection are $\gamma_{1} = 48, \gamma_{2} = 48, \gamma_{3} = 52, \gamma_{4} = 60$ . The detailed explanation on the process and experimental setup we used to determine the optimal $\gamma$ values are provided in Supp~\ref{supp:long-exposure exp}.

\begin{table*}[!ht]
\centering
\caption{Summary of the LENVIZ Training Dataset.} \label{tab:dataset_numbers} \vspace{-0.8em} 
 \small
\begin{tabularx}{0.8\textwidth}{r|X|X|X|X|X}
\hline
\multirow{2}{*}{\textbf{Camera Module}} & \multirow{2}{*}{\textbf{Resolution}} & \multicolumn{2}{c|}{\textbf{Human GT}} & \multicolumn{2}{c}{\textbf{Long exposure GT}} \\ \cline{3-6} 

 & & \textbf{\# of files} & \textbf{\# of scenes} & \textbf{\# of files} & \textbf{\# of scenes} \\ \hline  
 
S5K4H7YX03-FGX9 & 3264x2448 & 81,099 & 7,487 & 72,947 & 7,862 \\ 
S5KJN1SQ03      & 4080x3072 & 39,972 & 4,009 & 22,132 & 3,023 \\ 
S5KJNS          & 4080x3072 & 17,250 & 1,571 & 1,288 & 130 \\ \hline  
\multicolumn{2}{c|}{\textbf{Total}} &  138,321 & 13,067 & 96,367 & 11,015 \\ \hline 
\end{tabularx} \vspace{-1em} 
\end{table*}

\subsection{Post Capture Pipeline}\vspace{-0.25em}

Existing benchmark datasets for low-light image enhancement are primarily composed of either RAW\cite{MIT5K, SID, VE-LOL} or JPEG\cite{UHDLOL2023, LOL, DarkZurisch} images. A contrast between the traditional camera module-to-gallery pipeline and ours is demonstrated in Fig.~\ref{fig:capture_annotation_pipeline}. While RAW images offer a more pristine representation of the captured scene, their use is often restricted to commercial devices that provide access to raw data. JPEG images, on the other hand, are typically gallery images that have undergone extensive camera ISP tuning and default post-processing (such as face beautification, noise reduction, over-exposure suppression, etc). This processing can significantly alter the original camera module data, leading to a loss of fine details. As a result, models trained on these datasets may struggle to learn how to deal with camera module/ISP-tuning specific noise and fail to generalize to changes in camera's ISP and post-processing.

LENVIZ addresses the limitations of existing datasets by providing a collection of JPEG images that are exclusively ISP-tuned, devoid of any default camera post-processing. This approach ensures that the data closely resembles the raw output of the camera module+ISP pipeline, capturing the inherent noise, over-exposure, and other artifacts that are characteristic of low-light conditions. By training enhancement models on these ISP-tuned-only input frames, we effectively leverage the expertise of state-of-the-art ISP processing while also enhancing learning efficiency. 

\vspace{-0.5em}\subsection{Human Curated Ground truth Generation}\vspace{-0.25em}

Our human ground truth comprises a set of $13,067$ scenes edited by a team of 7 expert photographers and editors. The process followed a standardized workflow and style guide. The editors used the long-exposure shot as the base for the ground truth, as it offered a high-brightness view of the scene with considerably less noise than the multi-exposure frames. They then corrected and restored details lost to over-exposure using the multi-exposure frames, and were instructed to match the mid-exposure frames' color tones, focusing on technical quality: brightness, contrast and noise reduction.  Each edited image was then submitted for quality control and reviewed by three expert camera scientists, who evaluated it against a checklist of image quality aspects including noise levels, sharpness, and the absence of common artifacts like ghosting or color fringing, in addition to assessing color fidelity. Only images that received unanimous approval across all metrics were designated as the final ``Human GT" for the dataset. This structured and peer-reviewed approach provides a reproducible target for algorithms to learn from and ensures a high level of consistency and quality, providing a robust benchmark for training and evaluating low-light enhancement models. While the final product represents a specific, expertly-curated style, the reproducibility and rigor of our process make it an effective and reliable target for algorithm development.

\vspace{-0.5em}\section{LENVIZ Properties}\vspace{-0.5em}

In the following sections we will detail the specifics on the LENVIZ data distribution across different luminance levels, capture scenarios and content classes. Table~\ref{tab:dataset_numbers} provides a breakdown of the number of frames, scenes, and human-generated ground-truth in our dataset. Additional analysis such as feature distribution is provided in Supp~\ref{supp:feature_embdd}

\vspace{-.5em}\subsection{Luminance conditions}\vspace{-.25em}
Illuminance level is one of the primary criteria for selecting low-light scenes as it provides a quantitative measure of the scene's overall illumination (crucial in low-light images). We ensured that LENVIZ encompasses a variety of lighting conditions, from dimly lit rooms to outdoor scenes illuminated by street-lights, and paid special interest to the extreme-low light ranges 
(details on the illuminance distribution of our data are provided in Supp~\ref{fig:lux_histogram_global}). Images captured at such low levels of illumination provide a more challenging problem as they are more severely impacted by noise, lack of details, and color information and greatly benefit from having high quality annotations to serve as a guide for deep learning models on how to address these issues. 


\vspace{-.5em}\subsection{Capture scenarios}\vspace{-.25em} \label{sec:capture_scenarios}

One of the key aspects of LENVIZ is the richness of data we capture, demonstrated by our scene coverage (see  Fig. \ref{fig:teaser}). We have captured indoor locations ranging from regular living/dwelling areas, specialized photography studios, gaming and entertainment rooms, conference auditoriums, and classroom setups among the most notable ones; and outdoor scenes containing both urban and rural locations showcasing a wide range of backgrounds like general city buildings and night lights, greenery, fountains and other water sources, etc.  We made sure to provide a variety of illuminations sources with different intensities, colors, and complexities. These manifest in the forms of LED lights and signage both indoor and outdoors, string lights, directional lights, natural light, and ambient illumination. 

\textbf{Indoor scenes}: To comprehensively evaluate low-light image enhancement algorithms, our dataset includes indoor scenes characterized by uneven lighting and mixed light sources. This encompasses scenes with contrasting levels of illumination, such as brightly lit areas juxtaposed with dark corners or hallways. Additionally, we have captured scenes featuring a combination of incandescent and fluorescent lighting, whose differing color temperatures can pose significant challenges for image processing. These provide a robust testing ground for assessing the performance and generalizability of enhancement algorithms.

\textbf{Outdoor scenes:} As a complement to our indoor scenes, our dataset also includes a variety of outdoor scenes designed to challenge the capabilities of low-light image enhancement algorithms. These scenes cover scenarios with high dynamic ranges, where there's a stark contrast between brightly lit areas and darker elements. We have also captured scenes with strong directional lighting, such as sunsets or streetlamps, which introduce pronounced shadows. These outdoor scenarios provide a complex set of challenges for training and evaluating the performance of enhancement algorithms in real-world conditions.

\vspace{-0.5em}\subsection{Content classes}\vspace{-.25em}
\begin{figure}[t]
   \centering
   \includegraphics[width=0.9\linewidth]{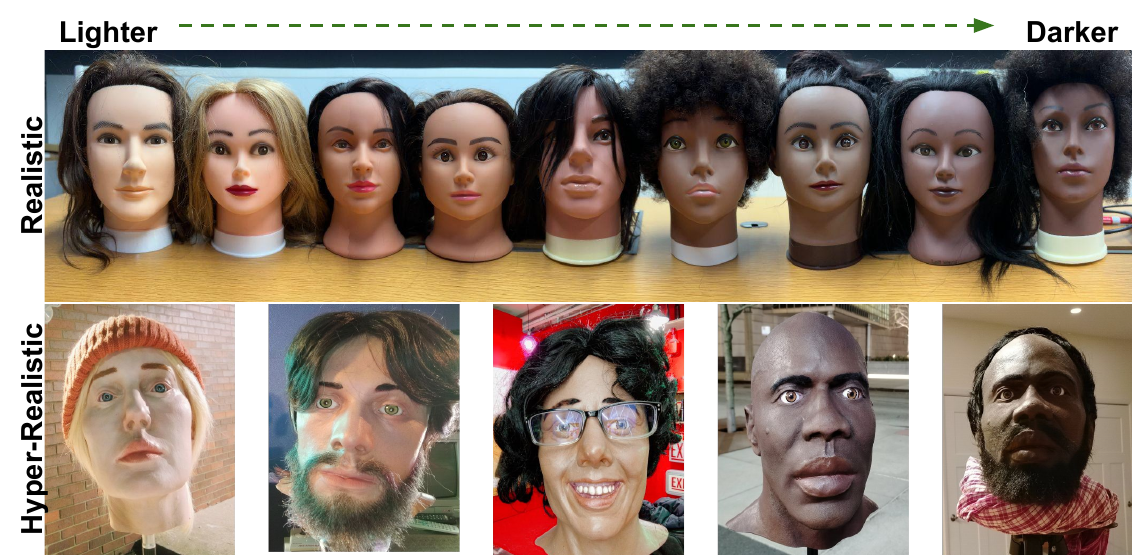} \vspace{-0.5em}
   \caption{Skin-tone variations and hyper-realistic mannequins. 
   }\label{mannequins}\vspace{-1.5em}
\end{figure} 


LENVIZ covers a broad spectrum of subjects and incorporates generous variations in scene content. We paid special attention to the presence of human and human-like subjects in our data, and to better encompass real-world conditions we incorporated a rich array of subject skin tones (14 mannequins spanning a diverse range of skin tones and facial features to ensure inclusivity), overall scene appearance (e.g., indoor vs. outdoor environments), and subject position (relative to the camera and light source). Since our long-exposure capture requires a higher restriction in motion, we limited the presence of humans to our test dataset, our train dataset contains 9 realistic and 4 hyper-realistic (Fig.~\ref{mannequins}) mannequins instead of humans for added control in the scenes. These were meticulously selected to exhibit natural variations in skin-tone, facial textures, pose, expression (simulated through head tilts and eye positioning), and accessories (such as glasses). This approach introduces a high degree of realism while maintaining control over lighting conditions and scene composition, ultimately leading to a robust and generalizable dataset for low-light image processing tasks. 
Overall, $70$\% of our scenes contain one or more faces, whose variation enhances the model's ability to learn from diverse camera usage scenarios.

To further characterize the dataset's scene content, we conducted a comprehensive object and face detection analysis using the AWS Label Detection\footnote{\url{https://docs.aws.amazon.com/rekognition/}} tool. This analysis served two primary purposes: 1) assessing the overall capability of commercial approaches to detect objects in extreme low-illuminance conditions, and 2) to identify unique features and content richness within each scene. The results of this analysis were incorporated as metadata for each of the scenes of our dataset, including detected bounding boxes, object classes (230 object labels belonging to 27 distinct categories), and in the case of face detection, face descriptors such as gender, presence of facial hair, or glasses. For a detailed breakdown on this please consult Supp.~\ref{supp:properties}.

\vspace{-.5em}\subsection{Test Dataset}\vspace{-0.25em} \label{test_set}

\begin{table}[th]
\centering
\vspace{-.5em}\caption{Summary of the LENVIZ Test Dataset.} \label{tab:test_dataset} \vspace{-0.5em} 
   \small
    \begin{tabular}{r|c|c} 
    \hline
    \textbf{Data-type} & \textbf{\# of files} & \textbf{\# of scenes} \\ \hline  
    Reference       & 610 & 60 \\ 
    No-reference    & 858 & 143\\ \hline  
    \textbf{Total}  & 1,468 & 203  \\ \hline 
    \end{tabular}  \vspace{-0.5em} 
\end{table}

To allow for an extensive evaluation of image quality in low-light conditions, our test dataset was designed to encompass a range of parameters, including sharpness, texture, contrast, brightness, naturalness, accurate skin-tone, over-exposure recovery, and more.  This resulted in a test dataset composed of 1,468 frames from 203 unique scenes. We provide two partitions: a ``Reference" partition including a paired human-edited ground truth and long-exposure shot, and a ``No-reference" partition which includes handheld captures and human subjects for a more challenging evaluation (see Table~\ref{tab:test_dataset} for details). Additional information on the test dataset such as the front and rear camera settings as well as test data examples is provided in Supp.~\ref{supp:test_data} 

While our test set is significantly smaller than our training data, its design prioritizes perceptual relevance over sheer volume. Having a high-quality, purpose-built test set offers superior insights than a larger less controlled one, which is critical for assessing low-light image enhancement algorithms where objective metrics often fall short of human perception. Drawing inspiration from the training dataset, we categorized scenes into two broad classes: indoor and outdoor. To account for diverse usage scenarios 
we included both selfie and non-selfie perspectives. Moreover, we incorporated human and non-human subjects in both indoor and outdoor settings to simulate real-world conditions. Furthermore, we adhered to DXO chart capture protocols for a subset of the scenes where we controlled the capture environment to facilitate the diagnostic analysis of specific image quality attributes such as noise reduction effectiveness, accurate color rendition, and dynamic range recovery under reproducible low-light conditions. This careful dataset curation ensures that the dataset thoroughly assesses the model's robustness and generalization across critical, perceptually distinct low-light scenarios.

\begin{figure*}[!ht]
   \includegraphics[width=0.97\linewidth]{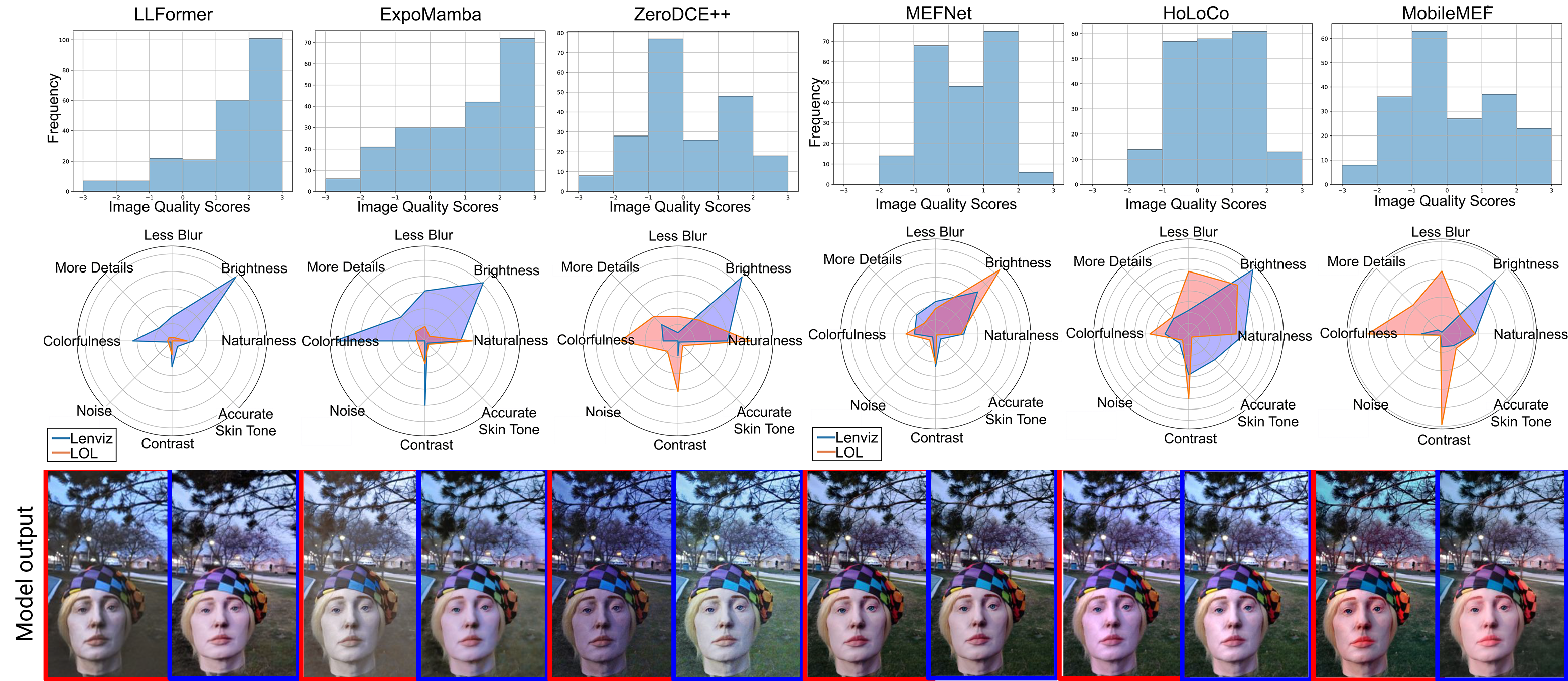}
   
   \caption{Comparison of model outputs trained on the LENVIZ and benchmark datasets. The histogram illustrates user preferences for model outputs from LENVIZ compared to others, while the spider chart highlights the advantages of various image quality aspects based on user votes.
   Left: LOL/SICE trained model output, Right: LENVIZ trained output}\vspace{-1em} 
   
   \label{fig:usertesting_all}
\end{figure*}

\vspace{-.5em}\section{LENVIZ Application}\vspace{-.5em}

To showcase the usability of our dataset for different low-light enhancement approaches, we provide an analysis on $3$ Single Exposure (LLFormer~\cite{UHDLOL2023}, ExpoMamba~\cite{adhikarla2024expomamba}, and ZeroDCE++~\cite{zero_dce++}) and $3$ Multi Exposure (MEFNet~\cite{mefnet}, HoLoCo~\cite{holoco}, and
MobileMEF~\cite{mobilemef}) low-light enhancement methods. Single exposure enhancement methods often use a low-exposure frame and attempt to improve its brightness, color, and detail recovery while reducing noise. Multi-exposure methods on the other hand incorporate the information of multiple frames captured at different exposure values, this often involves the use of a low exposure frame to recover information of over-exposed regions, and a medium or high exposure frame to get information related to color and details in under-exposed areas of the scene. All the $6$ approaches were trained using the LENVIZ dataset, as well as state-of-the-art single and multi-exposure benchmark datasets: LOL~\cite{LOL}, and SICE~\cite{SICE} respectively. 

\vspace{-.5em}\subsection{Quantitative Analysis}\vspace{-.25em}

While traditional objective metrics like PSNR are useful for quantitative reference-less assessment, they often fail to capture the perceived image quality from a human perspective, especially in low-light scenarios with complex noise and structural degradation. Taking this into account, we focused on perceptual metrics such as LPIPS and SSIM as a supplement, as these are more closely aligned with human perception. Our analysis revealed that models trained on our data consistently achieved superior scores in these metrics, demonstrating that our dataset is highly effective at training models to produce visually pleasing and structurally sound results. Furthermore, our cross-dataset evaluations confirm that these models generalize well to images from unseen cameras modules (more details on this in Supp~\ref{generalizability}). In contrast, PSNR results were more mixed, as this pixel-by-pixel metric is highly sensitive to subtle differences that might not be otherwise perceived by the human eye. A more detailed quantitative analysis, including a full table of results and cross-dataset performance, is provided in Supp~\ref{suppl:quantitative}. For in-depth understanding of the cases where LENVIZ trained model output we have added the analysis on failed cases in Supp~\ref{failed_case}.

\vspace{-0.5em}\subsection{Qualitative Analysis}\vspace{-.25em}\label{supp:qualitative}

Taking into account the limitations of quantitative metrics, we performed a comprehensive qualitative evaluation through an extensive user study (pairwise human evaluation (A/B testing)). This study provides direct insights into human perception of the enhancement quality prowess of the evaluated approaches, offering a more robust assessment in terms of visual fidelity, naturalness, and clarity under challenging illuminations. During the study, each participant was shown pairwise contrasting outputs from the LENVIZ and benchmark-trained models, participants were then asked to express their preference between the pairwise outputs and provide the primary reason for their choice. For a comprehensive evaluation, we selected 10 diverse scenes containing various objects, landscapes, illuminance levels, and both human subjects and mannequins. 
A total of 238 participants were randomly chosen from a pool of 89,883 Prolific candidates\footnote{\url{https://www.prolific.com/}}, ensuring diverse perspectives and unbiased feedback on model performance.


The histograms in Fig.~\ref{fig:usertesting_all} show the user satisfaction based on preference votes. Positive quality scores indicate a higher preference for images produced by models trained on the LENVIZ dataset. The y-axis represents the frequency of votes. Notably, models such as LLFormer~\cite{UHDLOL2023} and ExpoMamba~\cite{adhikarla2024expomamba} showed a significant preference for LENVIZ-trained outputs, for instance, with 84.44\% of participants favoring ExpoMamba's LENVIZ-trained images over ExpoMamba trained on LOL. In contrast, models like ZeroDCE++~\cite{zero_dce++}, MEFNet~\cite{mefnet}, and HoLoCo~\cite{holoco} performed comparably on both datasets. MobileMEF~\cite{mobilemef}'s performance on LENVIZ was slightly inferior, reflecting a potential gap between the model's performance and user expectations, despite this, when trained on LENVIZ data the user's highlighted overall better brightness as opposed to SICE-trained MobileMEF. It is important to note that 
the models were trained on an NVIDIA A100 SXM4 40 GB for 100 epochs, which may not have allowed full convergence.

Further analysis, as presented in the spider chart in Fig.~\ref{fig:usertesting_all}, highlights the key features affecting image quality that concerned users across different models. The majority of participants identified brightness as the most critical factor for favoring the LENVIZ-trained model outputs, followed by naturalness, colorfulness, and image detail.

\vspace{-0.5em}\section{Conclusion}\vspace{-0.5em}

In this paper, we introduced the Low Exposure Night Vision (LENVIZ) dataset, an industrial-grade resource designed to advance research in low-light imaging and encourage the community to address real-world challenges in low-light image and video processing. \textbf{LENVIZ comprises 24K paired real-world scenes}, both indoor and outdoor, with and without human presence. All images were captured under natural low-light conditions, featuring a wide range of lighting, noise levels, and scene complexities to reflect the true characteristics of low-light environments.

Using LENVIZ, we evaluated the performance of SOTA 
deep learning-based methods as well as their readiness for production-level implementation. 
Recognizing that traditional image quality metrics may not fully capture the subjective human experience, we also carried out extensive qualitative evaluations through end-user testing. The overall results 
suggested that the evaluated algorithms~\cite{UHDLOL2023, adhikarla2024expomamba, adhikarla2024unified, mefnet, holoco} show promise for real-world deployment, with further optimization enabled by our real-world data. We identified three key image quality aspects as the driving source of user preference, namely brightness, naturalness, and colorfulness. We hope that the LENVIZ dataset will serve as a valuable resource for future research, whether in advancing the understanding of human vision in low-light conditions or improving the performance of practical applications. 

\vspace{-0.5em}\section{Acknowledgment} \vspace{-0.5em} 
We thank Emily Quezada, Alicia Gonzalez, Jonathan San, and Damon Boler for their tireless efforts in collecting and editing our multi-sensor dataset scenes.
{
    \small
    \bibliographystyle{ieeenat_fullname}
    \bibliography{main}
}

\clearpage
\setcounter{equation}{0}
\setcounter{figure}{0}
\setcounter{table}{0}
\setcounter{page}{1}
\section*{Supplementary Material}\label{ref:Supp}
\subsection{Lenviz Capture}\label{supp:captures}

As mentioned in our the paper, the LENVIZ dataset has been captured using 3 camera module. While our S5K4H7YX03-FGX9 camera is better suited for close range captures, S5KJN1SQ03 and S5KJNS camera features (showcased in Table~\ref{tab:sensor_info}) allow for richer capture of details at medium and long range.

\subsubsection{Long-Exposure Time Calculation}\label{supp:long-exposure exp}

\begin{figure}[!h]
   \centering
   \includegraphics[width=\linewidth]{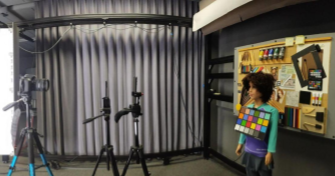}
   \caption{Long-exposure Illuminance time experimental setup. During the captures the room lighting was controlled and all the background lights were turned off to maintain the darkroom (low-light conditions).}  
   \label{fig:supp_eq2_p1}
\end{figure}

To empirically determine the values for the long-exposure parameters, a controlled experiment was conducted in a darkroom environment. This setup was designed to systematically measure and calibrate the relationship between scene illuminance and camera-captured image quality.

\textbf{Experimental Setup}
All experiments were performed in a controlled darkroom to eliminate external light sources and ensure precise illuminance levels. A high precision lux meter was used to accurately measure the illuminance of the scene at the subject's position, providing ground-truth illuminance values for calibration. For test subject, a standardized test chart (e.g., DXO standard chart along with deadleaves chart) and a a mannequin with a color palette were used, as indicated in the Fig~\ref{fig:supp_eq2_p1}. These provided a consistent reference for evaluating sharpness, texture and color fidelity the image quality that from our study was highly effected in low-light conditions. Regarding camera configurations, the phones with camera modules listed in Table~\ref{tab:sensor_info} was mounted on a tripod to ensure stability and was configured to capture long-exposure images. Simultaneously, a professional grade DSLR camera (Canon EOS R6 Mark II Mirrorles \& Canon RF 15-35mm f/2.8 L IS USM Lens) was placed on a separate tripod with its field of view overlapping the phone's. The DSLR was set to auto exposure mode to provide a consistent, high-quality reference for image metrics.

\begin{figure}[!h]
   \centering
   \includegraphics[width=\linewidth]{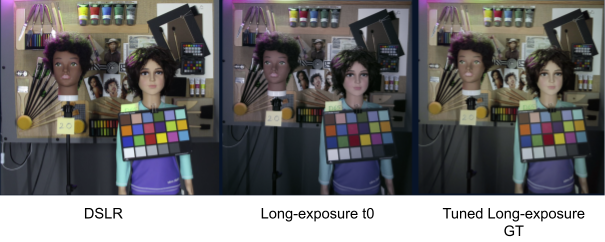}
   \caption{Examples of images captured during the empirical tuning of long-exposure time. Left image is DSLR captured as the control at lux meter reading=2 (post in note in scene indicates the lux reading of scene), middle image represents the long-exposure capture at exposure setting of 4 seconds, right image represents the final tuned long-exposure captured at time based on Eq~\ref{equation 2}.}  
   \label{fig:supp_eq2_p2}
\end{figure}

\textbf{Experimental Procedure}
First, for illuminance calibration, a lux-meter was used to measure and verify the illuminance levels for each scene. The lights in the darkroom were adjusted to create a series of controlled illuminance settings raging from 0.1 lux to 50 lux, with 1 lux increments as measure by external lux-meter. During image capture, images were captured simultaneously with both the phone and the DSLR at each illuminance level. This process was repeated for long-exposure with the phone's shutter speed ranging from 2 seconds to 30 seconds in 2 second increments resulting in 15 sets (each containing scenes between lux 0.1 to 50). The DSLR images was captured on the first set. So the total of 16 sets (15 from phone and 1 from DSLR) were then submitted to DXOMark (professional imaging lab) for analysis. The lab conducted a detailed evaluation of key camera tests, including Modulation Transfer Function (MTF), deadleaves chart analysis, and noise characteristics. The final parameter tuning for long-exposure was based on the reports of camera testing from DXOMark, along with expert visual inspection against DSLR reference images, were used to tune the gamma parameter of Eq~\ref{equation 2}. These parameters were adjusted to the closest possible match in terms of sharpness, texture, and color fidelity between the phone's camera long-exposure output and the professional DSLR's reference images based on the test results for each images from DXOMARK . The example of the DSLR reference image as well as before and after tuned GT output is shown in Fig~\ref{fig:supp_eq2_p2}.

\subsubsection{Luminance VS Exposure time} \label{fig:lux_est_image}

The estimated exposure time for our shots is plotted against the estimated illuminance, as the estimated scene illuminance increases, the exposure time reduces accordingly to minimize the occurrence of over-exposed regions. 
\begin{figure}[h]
   \centering
   \vspace{-1em}
   \includegraphics[width=\linewidth]{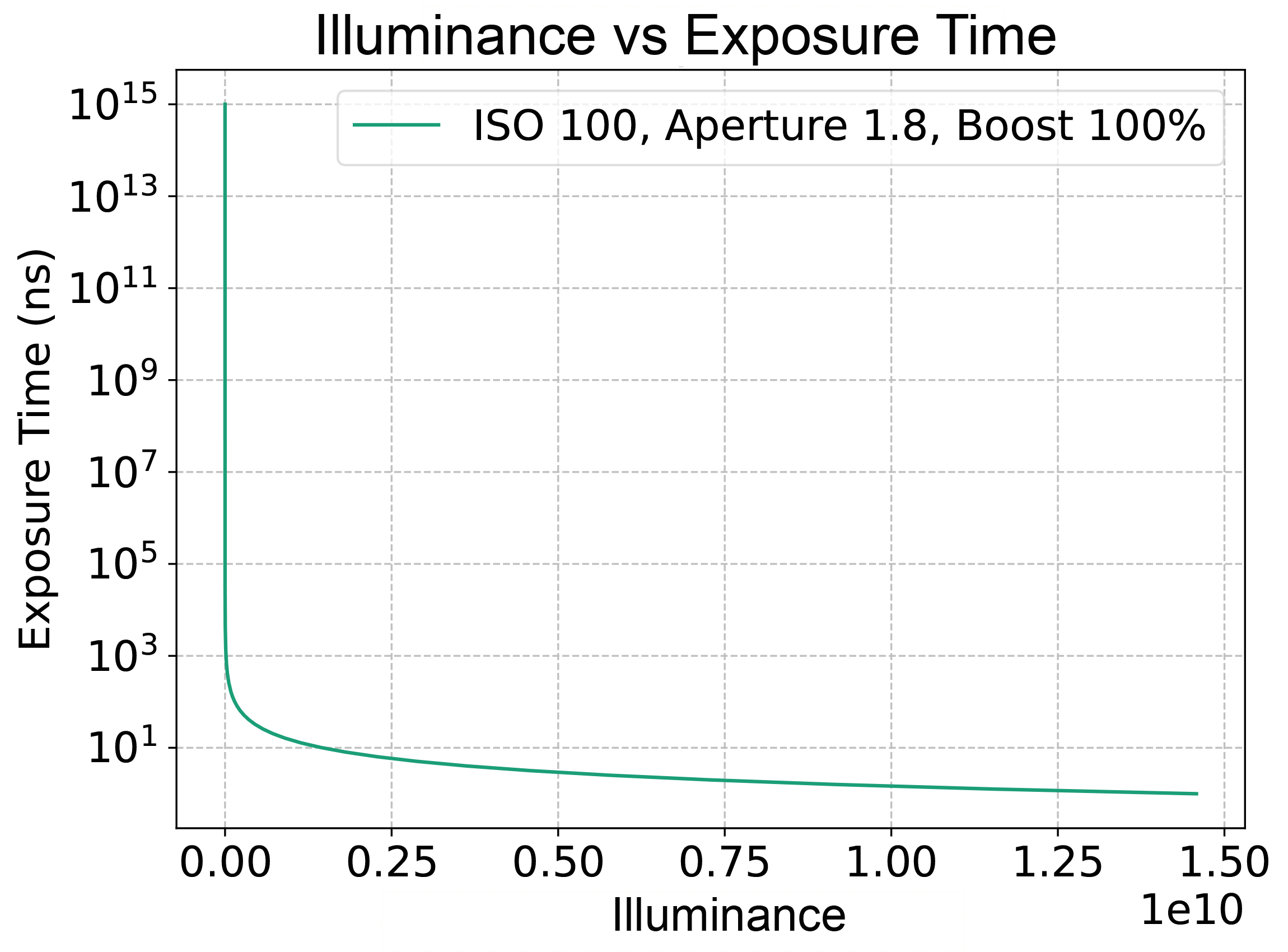}
   \includegraphics[width=\linewidth]{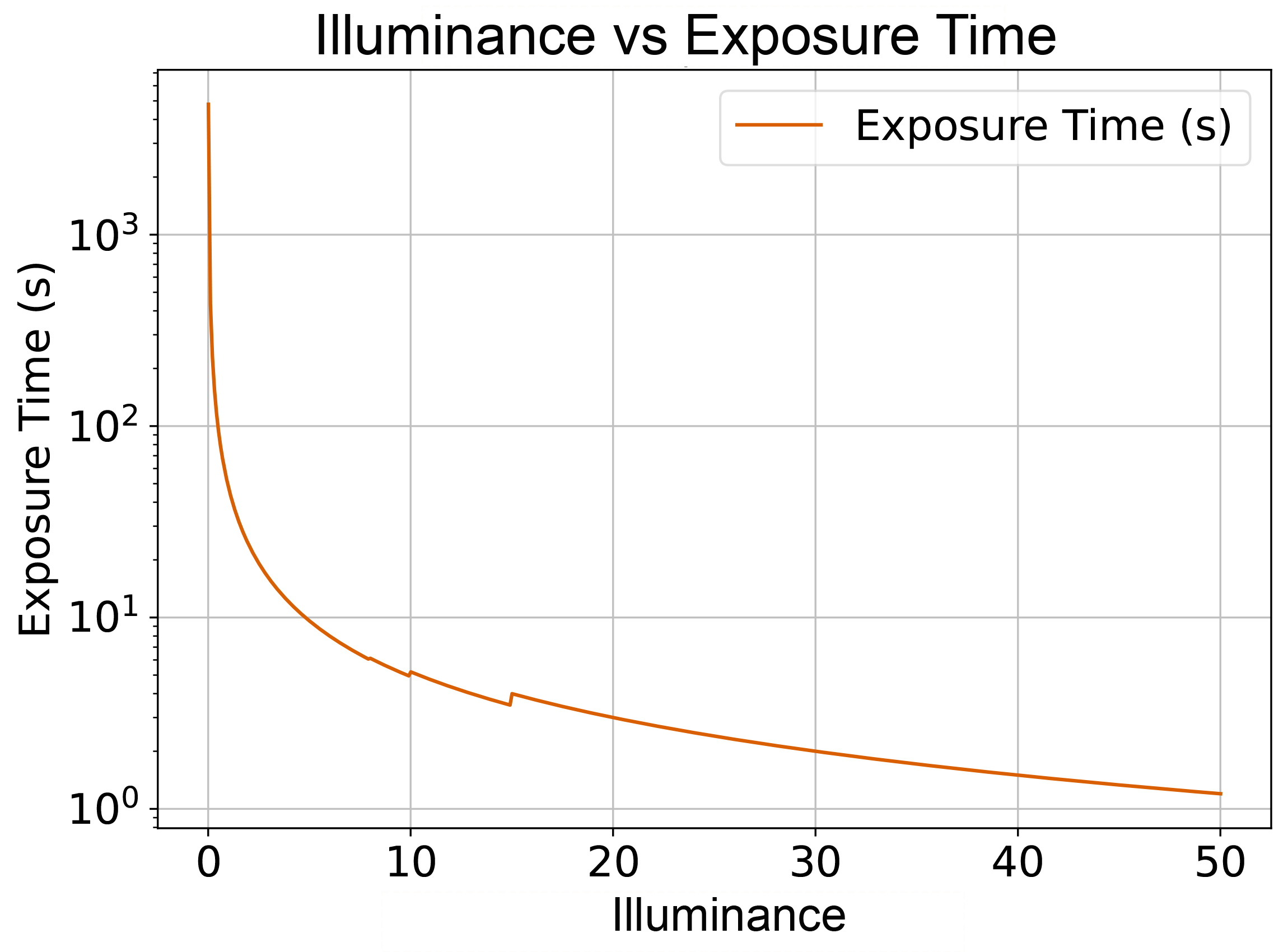}\vspace{-1em}
   \caption{Illuminance vs exposure time (log scale) [\textbf{Top}]~ For standard values of ISO 100, aperture 1.8 and no additional boost. [\textbf{Bottom}]~For long-exposure shot.}\vspace{-1em}
\end{figure}

\subsection{LENVIZ Additional Properties}\label{supp:properties}

\begin{table*}[h]
\caption{Camera Module Specifications}\label{tab:sensor_info}
\resizebox{\textwidth}{!}{
\begin{tabular}{|l|c|c|c|c|c|c|}
\hline
\textbf{Camera module Name }    & \begin{tabular}[c]{@{}c@{}}\textbf{Resolution} \\ \textbf{(MP)}\end{tabular} & \textbf{Aperture} & \begin{tabular}[c]{@{}c@{}}\textbf{Pixel-size}\\ \textbf{(um)}\end{tabular} & \textbf{Camera-Size} & \textbf{Focus}      & \textbf{FOV (Diag)} \\ \hline \hline
S5K4H7YX03-FGX9 & 8  & f/2.0    & 1.12   & 1/3         & Fixed (27cm$\sim$39.3cm) & 78°        \\ \hline
S5KJN1SQ03      & 50 & f/1.8    & 0.64   & 1/2.76"     & Auto (10cm$\sim$INF)     & 74.26°     \\ \hline
S5KJNS & 50 & f/1.8    & 0.64   & 1/2.76"     & Auto (10cm$\sim$INF)     & 74.26°     \\ \hline
\end{tabular}}
\end{table*}

\subsubsection{Content class}\label{supp:content_class}
\begin{figure}[h]
   \centering
   \includegraphics[width=1\linewidth]{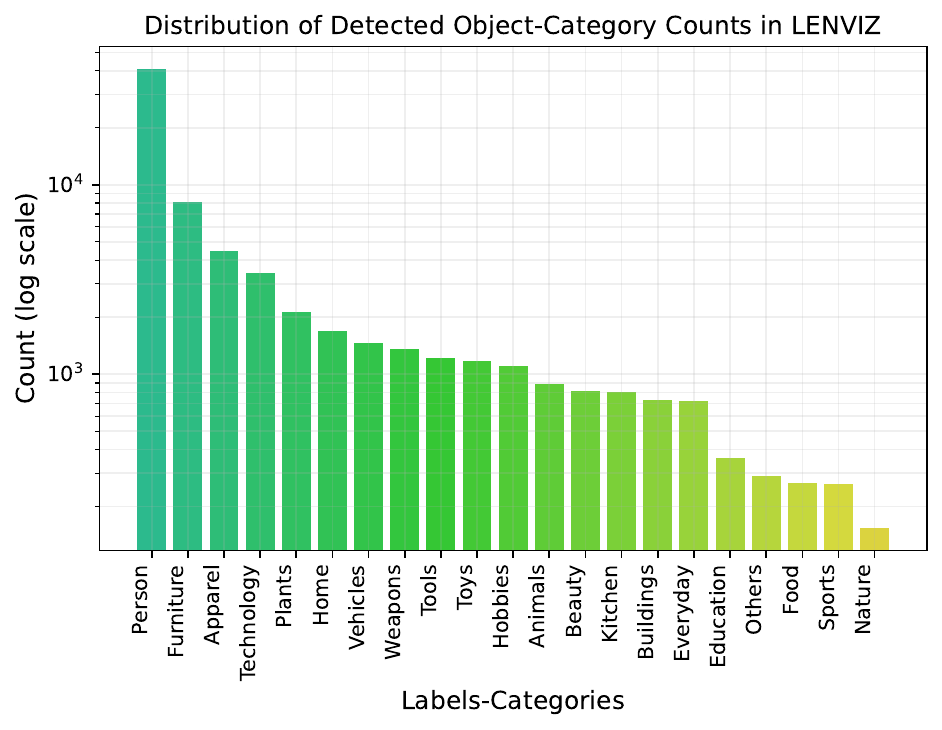}
   \caption{Demonstration of \textbf{27 unique object categories} comprising the \textbf{230 object labels} throughout the LENVIZ dataset.}\label{fig:object-labels} 
\end{figure} 

Fig~\ref{fig:object-labels} provides an overlook at the distribution of the different object categories identified by the AWS object detection algorithm. In general, the detected objects within our images can be classified into 27 broad categories, ranging from "Plants" and "Animals" to "Buildings" and "Vehicles". Each of these categories are further divided into 230 object-specific labels like "Chair" or "Couch" for the general "Furniture" Category. Section~\ref{supp:properties} Table~\ref{tab:object_breakdown} provides a breakdown of each of the 230 object labels and their respective categories. We include the list of object labels identified for each given scene as well as their bounding boxes in our dataset release.
\subsubsection{Illuminance distribution}\label{fig:lux_histogram_global}
\begin{figure}[h]
   \centering
   \includegraphics[width=0.8\linewidth]{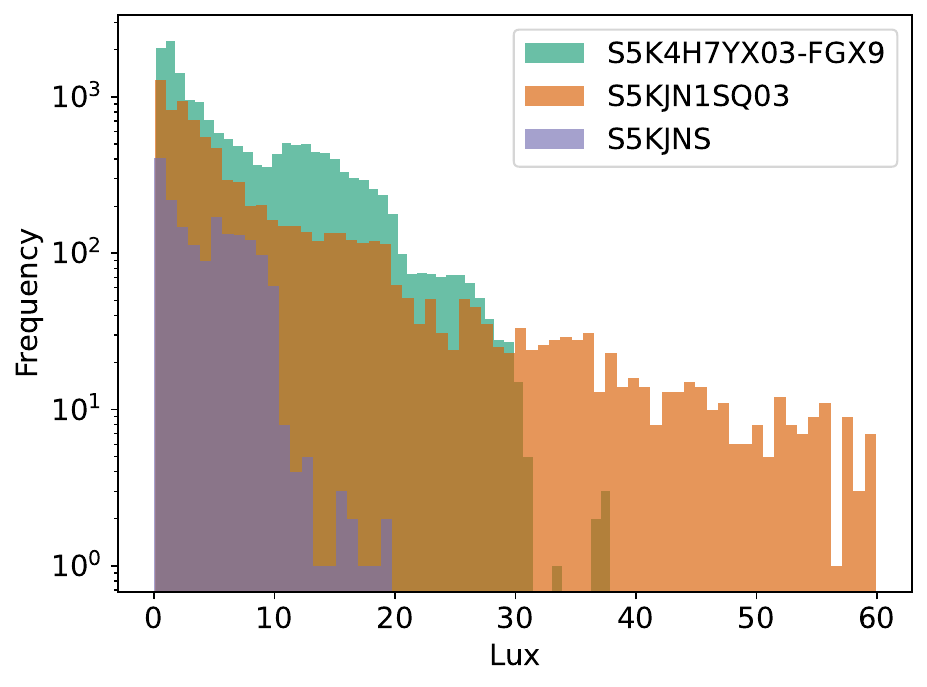} \vspace{-1em}
   \caption{LENVIZ Illuminance Histogram} \vspace{-1em}
\end{figure}

\subsubsection{Feature Embedding Analysis}\label{supp:feature_embdd}

\begin{figure}[!h]
   \centering
   \includegraphics[width=\linewidth]{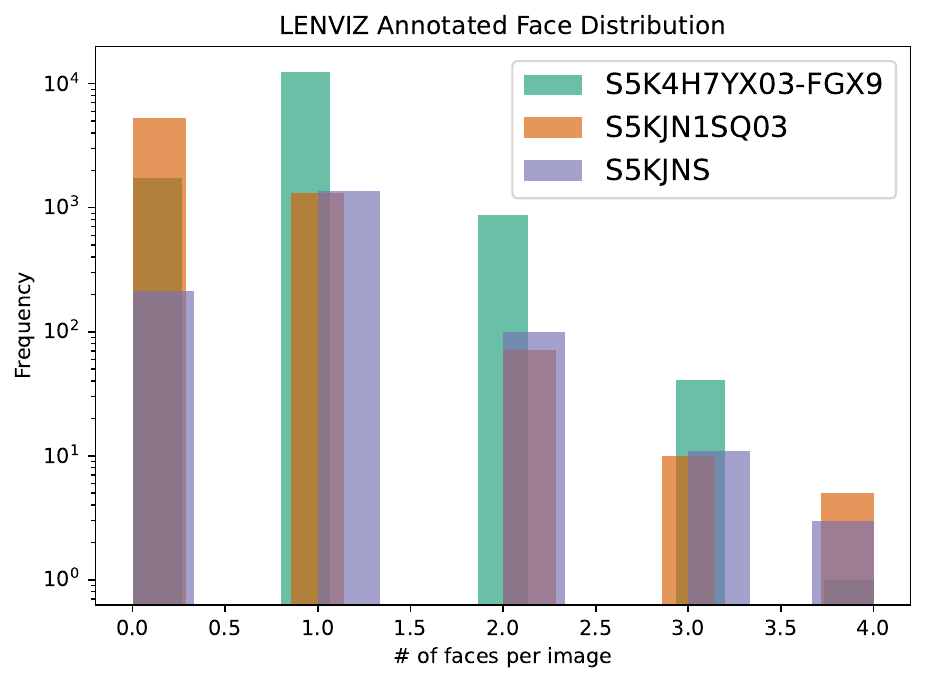}
   \caption{Number of Faces Per Scene Histogram}  
   \label{fig:supp_lux_histogram_faces}
\end{figure}

\begin{figure}[!ht]
   \centering
   \includegraphics[width=1\linewidth]{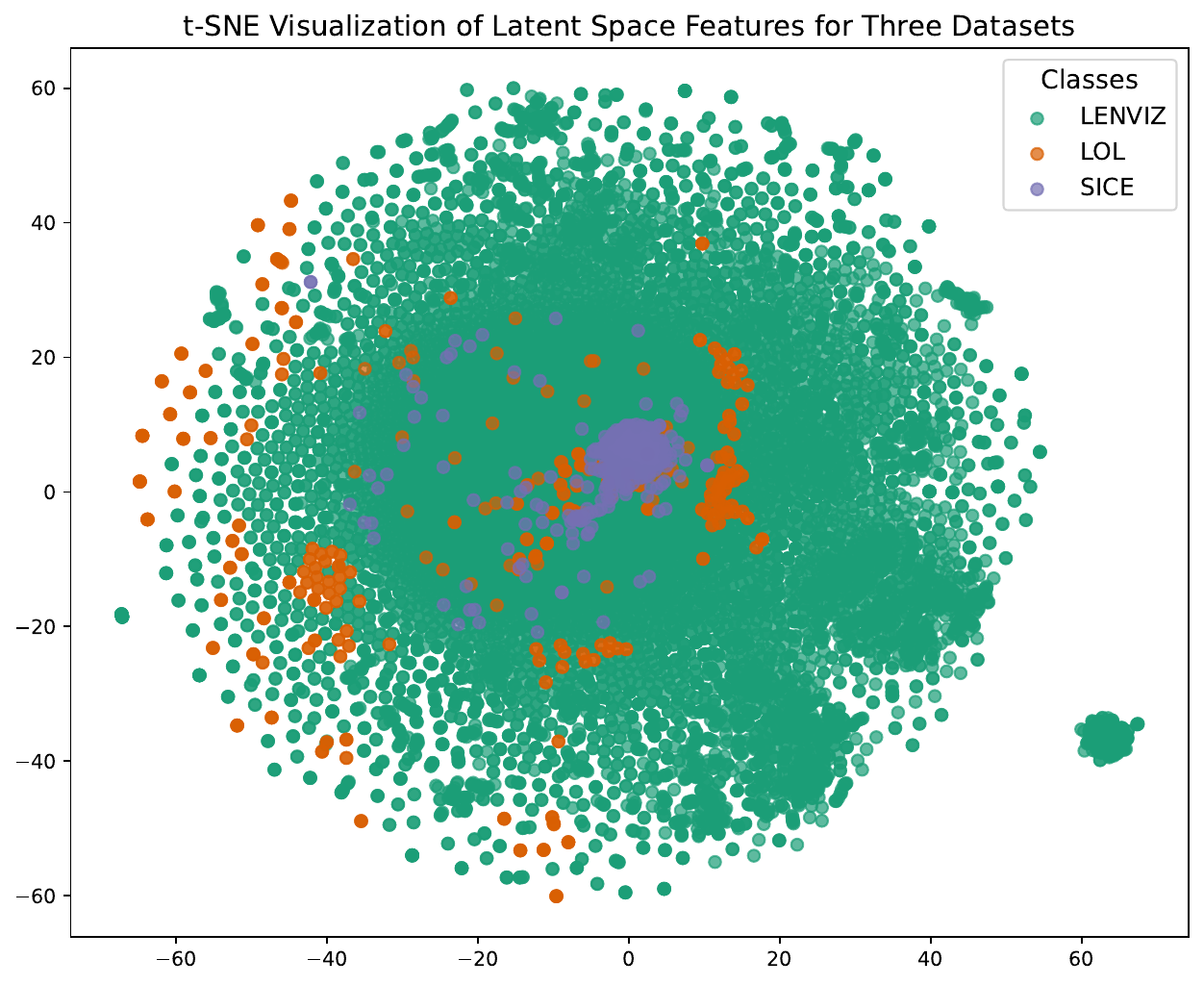}
   \caption{Feature Embedding comparison of LENVIZ Long-exposure frames vs LOL vs SICE}\label{fig:feature_embb_1} 
\end{figure} 

\begin{table*}[ht]
\caption{Object Categories and Labels Breakdown}\label{tab:object_breakdown}
\resizebox{\textwidth}{!}{
\begin{tabularx}{\linewidth}{|l|l|X|}
\hline
    \textbf{Category} & \textbf{Objects} & \textbf{Object labels} \\ \hline
    Person Description & 12 & Girl, Man, Boy, Person, Baby, Female, Adult, Male, Child, Woman, Teen, Bride \\ \hline
    Plants and Flowers & 3 & Plant, Rose, Fungus \\ \hline
    Technology and Computing & 14 & Laptop, Disk, Speaker, Microphone, VR Headset, Mobile Phone, Monitor, Earbuds, Remote Control, QR Code, Computer Keyboard, Camera, Headphones, Tablet Computer \\ \hline
    Toys and Gaming & 1 & Doll \\ \hline
    Furniture and Furnishings & 12 &  Chandelier, Ceiling Fan, Chair, Rug, Bench, Dining Table, Door, Photo Frame, Lamp, Desk, Painting, Couch\\ \hline
    Beauty and Personal Care & 3 &  Toothbrush, Lipstick, Tattoo\\ \hline
    Kitchen and Dining & 5 & Fork, Plate, Shaker, Spoon, Cup \\ \hline
    Buildings and Architecture & 5 & Windmill, Tower, Clock Tower, Building, Gate \\ \hline
    Tools and Machinery & 11 & Power Drill, Hammer, Baton, Switch, Brush, Blow Dryer, Screwdriver, Scissors, Tape, Shovel, Screw \\ \hline
    Apparel and Accessories & 32 & Shirt, Shorts, Bridal Veil, Box, High Heel, Glasses, Hat, Wristwatch, Sunglasses, Sweater, Overcoat, Coat, Suit, Wallet, Tie, Glove, Diamond, Handbag, Belt, Bracelet, Shoe, Ring, Necklace, Razor, Sock, Helmet, Locket, Perfume, Backpack, Jacket, Jeans, Scarf \\ \hline
    Home and Indoors & 19 & Swimming Pool, Hot Tub, Sink Faucet, Crib, Staircase, Fireplace, Package, Mailbox, Bathtub, Toilet, Bed, Sink, Shower Faucet, Mixer, Lawn Mower, Washer, Cooktop, Refrigerator, Microwave  \\ \hline
    Weapons and Military & 13 & Dagger, Mace Club, Spear, Axe, Bow, Mortar Shell, Gun, Crossbow, Dynamite, Sword, Grenade, Arrow, Knife \\ \hline
    Vehicles and Automotive & 11 & Wheel, Boat, Bus, Airplane, Gas Pump, Pickup Truck, Train, Truck, E-scooter, Motorcycle, Car \\ \hline
    Food and Beverage & 16 & Pear, Egg, Lobster, Milk, Burger, Beer, Orange, Ice Cream, Bread, Apple, Hot Dog, Banana, Pineapple, Can, Ketchup, Pizza \\ \hline
    Hobbies and Interests & 9 & Toy, Bicycle, Violin, Clapperboard, Piano, Book, Teddy Bear, Guitar, Smoke Pipe \\ \hline
    Nature and Outdoors & 1 & Moon \\ \hline
    Symbols and Flags & 2 & Flag, Cross \\ \hline
    Sports & 18 &  Field Hockey Stick, Ice Hockey Puck, Rugby Ball, Ping Pong Paddle, Baseball (Ball), Soccer Ball, Baseball Glove, Volleyball (Ball), Cricket Bat, Scoreboard, Ice Hockey Stick, Skateboard, Baseball Bat, Tennis Ball, Basketball (Ball), American Football (Ball), Chess, Cricket Ball \\ \hline
    Animals and Pets & 25 & Dinosaur, Honey Bee, Dog, Spider, Insect, Turtle, Fish, Giraffe, Kangaroo, Mouse, Lion, Chicken, Antelope, Elephant, Cat, Bird, Tiger, Bear, Sheep, Lizard, Horse, Shark, Snake, Pig, Zebra \\ \hline
    Education & 1 & Blackboard \\ \hline
    Text and Documents & 4 & Business Card, Credit Card, Passport, Driving License \\ \hline
    Everyday Objects & 3 & Disposable Cup, Candle, Pen \\ \hline
    Offices and Workspaces & 1 & White Board \\ \hline
    Transport and Logistics & 3 & Traffic Light, Road Sign, Utility Pole \\ \hline
    Events and Attractions & 3 & Hanukkah Menorah, Balloon, Snowman \\ \hline
    Medical & 2 & First Aid, Pill\\ \hline
    Public Safety & 1 & Fire Hydrant \\ \hline
\end{tabularx}
}
\end{table*}

To understand the underlying structure and relationships within our LENVIZ dataset, we analyzed the feature distribution of our scenes in a latent space representation and compared it against other benchmakring datasets in the field, namely, LOL~\cite{LOL} (Single-Exposure), and SICE~\cite{SICE} (Multi-Exposure). We extracted a deep feature representation for each of the scenes of all three datasets using the output of the last convolutional block in the VGG16 model given its well known suitability for feature extraction and its ability to recognize a vast range of visual patterns. To reduce the dimensionality of these features and visualize their distribution, we applied t-SNE to map these high-dimensional data points to a lower-dimensional space while preserving their local structure. As seen in Fig~\ref{fig:feature_embb_1}, the t-SNE visualization reveals that both LOL and SICE have some degree of overlap. The distribution of features from the LOL datasets seems to be relatively widespread in comparison to SICE. Despite this, our dataset is able to encompass a much wider feature distribution than these two benchmark datasets, as such allowing for models to learn from a wider set of features when using LENVIZ.

\subsubsection{Test Dataset}\label{supp:test_data}

\begin{figure*}[!h]
   \centering
   \includegraphics[width=1\linewidth]{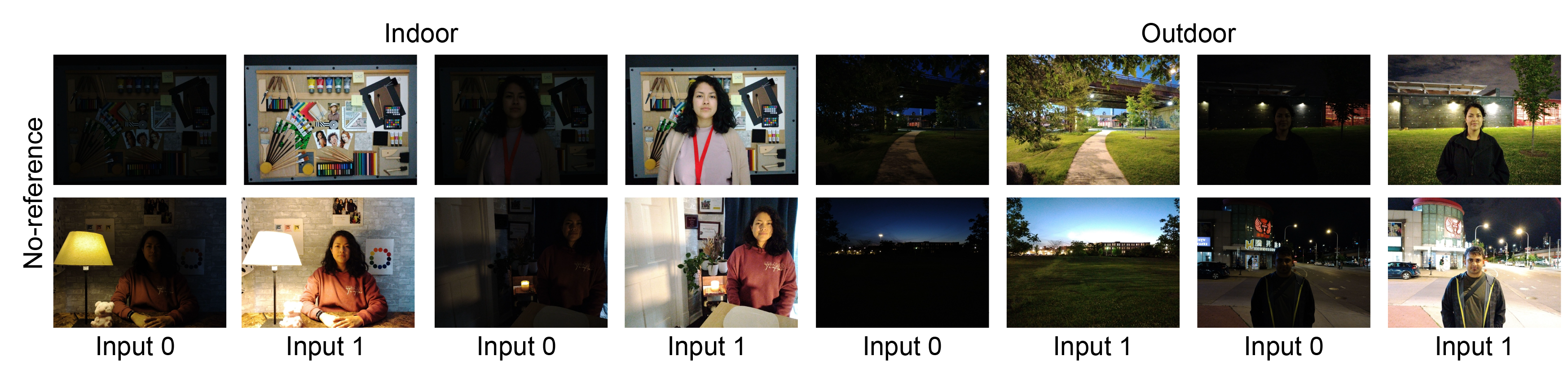}
   \includegraphics[width=1\linewidth]{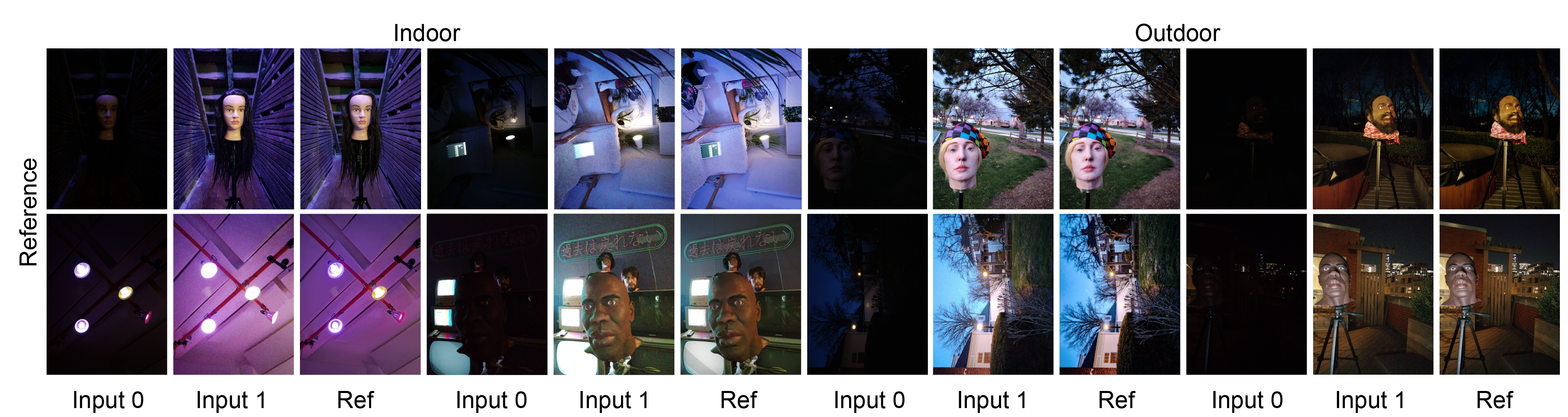}
   \vspace{-2em} \caption{Example of LENVIZ No-Reference [Top] and Reference [Bottom] Test data}  \vspace{-1em} 
   \label{fig:supp_test_data}
\end{figure*}

To introduce illumination variations, we systematically adjusted illuminance levels from 0 to 20 for rear imagers (S5KJN1SQ03 \& S5KJNS) and from 0 to 30 for front imagers (S5K4H7YX03-FGX9). Furthermore, we included both flash-on and flash-off conditions to replicate authentic low-light scenarios encountered in everyday photography.

For a comprehensive evaluation of background influence and model robustness, we  analyzed images captured by renowned testing platforms like DXOmark\footnote{\url{https://corp.dxomark.com/}} and our internal units. Drawing inspiration from these sources, we incorporated diverse background elements into our test dataset, including DXOmark charts, solid-colored wallpaper, polka dots (for ringing artifact analysis), text, and more. With all these variations we are additionally providing a reference (human-generated low-light enhancement ground truth) as well as no-reference types of test dataset. This is helpful to initially evaluate the model IQ using no-reference dataset and later use the reference as stage two evaluation quantitatively and qualitatively. Fig~\ref{fig:supp_test_data} illustrates representative test scenes for both indoor and outdoor environments.

Our test dataset distinguishes itself from existing benchmarks not only by its broader range of scene types but also by its emphasis on model robustness. These meticulously designed scenes are intended to challenge the trained enhancement models, assessing their ability to maintain consistent performance under varying low-light conditions and ensuring stable reproducibility.

\subsection{Lenviz Additional Application}
\subsubsection{SOTA Models} \label{supp:quant_analysis}
\begin{table*}[t]
\begin{center}
\caption{Quantitative Evaluation results for SOTA single exposure (SE) methods and Multi-exposure fusion (ME) methods on LENVIZ dataset. Here, higher the PSNR and SSIM value and lower the LPIPS score indicated that better the output quality }\label{tb:sota_quant_metrics}
\begin{tabular}{l|l|l|ll|ll|ll}
\hline
\multirow{2}{*}{\textbf{Type}} & \multirow{2}{*}{\textbf{SOTA}} & \multirow{2}{*}{\textbf{Test Dataset}} & \multicolumn{2}{c|}{\textbf{\textcolor{green}{$\uparrow$}~PSNR}}                                                                                                           & \multicolumn{2}{c|}{\textbf{\textcolor{green}{$\uparrow$}~SSIM}}                                                                                                           & \multicolumn{2}{c}{\textbf{\textcolor{green}{$\downarrow$}~LPIPS}}                                                                                                          \\ \cline{4-9} 
                               &                                &                                        & \multicolumn{1}{l|}{\begin{tabular}[c]{@{}l@{}}LENVIZ \\ trained\end{tabular}} & \begin{tabular}[c]{@{}l@{}}LOL/SICE \\ trained\end{tabular} & \multicolumn{1}{l|}{\begin{tabular}[c]{@{}l@{}}LENVIZ\\  trained\end{tabular}} & \begin{tabular}[c]{@{}l@{}}LOL/SICE \\ trained\end{tabular} & \multicolumn{1}{l|}{\begin{tabular}[c]{@{}l@{}}LENVIZ \\ trained\end{tabular}} & \begin{tabular}[c]{@{}l@{}}LOL/SICE \\ trained\end{tabular} \\ \hline

\multirow{8}{*}{SE}   & \multirow{2}{*}{ZeroDCE++~\cite{zero_dce++}} & LENVIZ                        & \multicolumn{1}{l|}{16.35}   &   \textbf{16.38}           & \multicolumn{1}{l|}{0.359}  &        \textbf{0.437 }     & \multicolumn{1}{l|}{0.663}     &   \textbf{0.611}        \\ 
                      &                            & LOL                           & \multicolumn{1}{l|}{14.75}               &      14.75            & \multicolumn{1}{l|}{0.518}               &       0.518         & \multicolumn{1}{l|}{0.328}               &         0.328         \\ 
                      & \multirow{2}{*}{LLFormer~\cite{UHDLOL2023}}  & LENVIZ                        & \multicolumn{1}{l|}{\textbf{21.13}} & 17.05 & \multicolumn{1}{l|}{\textbf{0.665}} & 0.551 & \multicolumn{1}{l|}{\textbf{0.358}} &  0.498               \\ 
                      &                            & LOL                           & \multicolumn{1}{l|}{19.33}               &  \textbf{19.79 }        & \multicolumn{1}{l|}{\textbf{0.757}}    &   0.772       & 
                      \multicolumn{1}{l|}{\textbf{0.240}}               &     0.277             \\ 
                      & \multirow{2}{*}{ExpoMamba~\cite{adhikarla2024expomamba}} & LENVIZ                        & \multicolumn{1}{l|}{\textbf{19.8 }} & 17.04 & \multicolumn{1}{l|}{\textbf{0.585}} & 0.534   & \multicolumn{1}{l|}{\textbf{0.524}}   &   0.599           \\ 
                      &                            & LOL                           & \multicolumn{1}{l|}{\textbf{23.65}}               &         18.55         & \multicolumn{1}{l|}{\textbf{0.816}}   &     0.759             & \multicolumn{1}{l|}{\textbf{0.169}}               &       0.291           \\ \hline
\multirow{8}{*}{ME}                      & \multirow{2}{*}{MEFNet~\cite{mefnet}}    & LENVIZ                        & \multicolumn{1}{l|}{20.71} & \textbf{20.77} & \multicolumn{1}{l|}{\textbf{0.609}} &   0.606   & \multicolumn{1}{l|}{\textbf{0.457}}               &    0.458    \\ 
                      &                            & SICE                          & \multicolumn{1}{l|}{\textbf{21.13}} &  20.94   & \multicolumn{1}{l|}{0.612}&  0.612  & \multicolumn{1}{l|}{\textbf{0.358}}               &      0.361            \\ 
                      & \multirow{2}{*}{HoLoCo~\cite{holoco}}    & LENVIZ                        & \multicolumn{1}{l|}{21.31} &  \textbf{20.77}      & \multicolumn{1}{l|}{\textbf{0.613}}  & 0.606 & \multicolumn{1}{l|}{0.689}               &     0.689             \\ 
                      &                            & SICE                          & \multicolumn{1}{l|}{13.78}  &  \textbf{13.90}  & \multicolumn{1}{l|}{0.614}& \textbf{0.615} & \multicolumn{1}{l|}{\textbf{0.526}} &   0.529               \\ 
                      & \multirow{2}{*}{MobileMEF~\cite{mobilemef}} & LENVIZ                        & \multicolumn{1}{l|}{\textbf{20.93}}               &       19.47           & \multicolumn{1}{l|}{\textbf{0.626}}               &       0.613           & \multicolumn{1}{l|}{\textbf{0.492}}               &       0.561           \\ 
                      &                            & SICE                          & \multicolumn{1}{l|}{13.65} &  \textbf{14.36}  & \multicolumn{1}{l|}{\textbf{0.637}}  &  0.632  & \multicolumn{1}{l|}{0.484}  &   \textbf{0.452}  \\ \hline
\end{tabular}
\end{center}
\end{table*}
\textbf{Single Exposure Approaches}: Zero-DCE++ \cite{zero_dce++}, leverages a zero-reference deep curve estimation technique to enhance images without the need for paired ground-truth data, focusing on real-time illumination adjustment to improve brightness and contrast. LLFormer \cite{wang2023ultra}, incorporates axis-based multi-head self-attention and cross-layer attention fusion. Finally, ExpoMamba \cite{adhikarla2024expomamba} introduces a novel architecture that integrates frequency state space components in a Mamba (a state space model family) to tackle real-time processing challenges. 

\textbf{Multi-Exposure Approaches}: 
MEFNet~\cite{mefnet} employs a multi-exposure fusion network that predicts the fusion weight maps at a low resolution for fast processing, HoLoCo~\cite{holoco} introduces contrastive learning with a holistic and local contrastive constraint to multi-exposure image fusion to recover details and allow for uniform illumination in over and under exposed regions. MobileMEF~\cite{mobilemef} is a lightweight multi-exposure fusion network for real-time processing on mobile platforms.

For training, we used our entire set of Human-edited Ground Truth Training data scenes ($13,067$ scenes total). All low-light methods were trained for 100 epochs using the original implementations and hyper-parameters provided by their respective authors. When training Single exposure methods, we used frames captured at a low exposure value (EV -20) as the input for the models. For Multi exposure approaches, we provided two frames as input, one captured at low exposure (EV -20), and one at medium exposure (EV 0). The selection of these exposure values was done to closely follow the observed illumination of the input frames provided in the LOL and SICE training data.

To measure the performance of these methods, we conducted a quantitative and qualitative evaluation to assess their image quality when trained with our dataset in contrast with the results obtained when training with existing benchmark datasets. We further evaluate the generalization capabilities of each approach by performing cross-dataset evaluation. For our quantitative evaluation we included evaluation metrics such as PSNR, SSIM, and LPIPS. Our qualitative evaluation consisted of a human study to evaluate the perceived image quality of each method's outputs. We incorporated the feedback of 238 users who rated the outputs and provided insights on their perceived quality based on eight features: naturalness, brightness, blur, details, colorfulness, noise, contrast, and skin tone accuracy.

\subsection{Quantitative evaluation}~\label{suppl:quantitative}

To complement our extensive user study and provide a comprehensive quantitative assessment, we evaluated the performance of six state-of-the-art low-light enhancement models when trained on our dataset and on two leading benchmark datasets (LOL and SICE). The results summarized in Table~\ref{tb:sota_quant_metrics}, highlight the superior performance of models trained on our data across key metrics.

The evaluation included three primary metrics: LPIPS (Learned Perceptual Image Patch Similarity), SSIM (Structural Similarity Index Measure) and PSNR (Peak Signal-to-Noise Ratio). LPIPS is a perceptual metric that uses a deep learning model to measure the similarity of two images as a human would perceive it. SSIM evaluates image quality based on structural information, brightness, and contrast, providing a more human-centric score than tradition pixel based metrics. PSNR, in contrast, is a simple pixel-by-pixel comparison that is highly sensitive to small pixel shifts or variations. 

Our findings reveal an overwhelming trend of superior performance in LPIPS and SSIM for models trained on our dataset, both on our test set and in cross-dataset evaluations. This indicates that our dataset is exceptionally effective at training models that produce perceptually pleasing images with high structural and aesthetic fidelity. With only minor exceptions, such as ZeroDCE++ and MobileMEF obtaining slightly better results when trained on LOL and SICE (~0.05 and 0.03 respectively). This performance is almost entirely mirrored in the SSIM metric, with LENVIZ-trained models consistently outperforming their counterparts, with slightly more variation observed among multi-exposure methods trained on SICE.

In contrast to the perceptual metrics, the results for PSNR were more mixed. While models trained on our dataset still secured superior scores in a significant portion of the test cases, the overall distribution of scores was less consistent across methods and datasets. This is because PSNR is a strict, pixel-by-pixel metric that is sensitive to subtle differences in color, brightness, and alignment that a human eye would not notice. 



    
\subsubsection{Failure cases analysis}\label{failed_case}
\begin{figure*}[!h]
   \centering
   \includegraphics[width=1\linewidth]{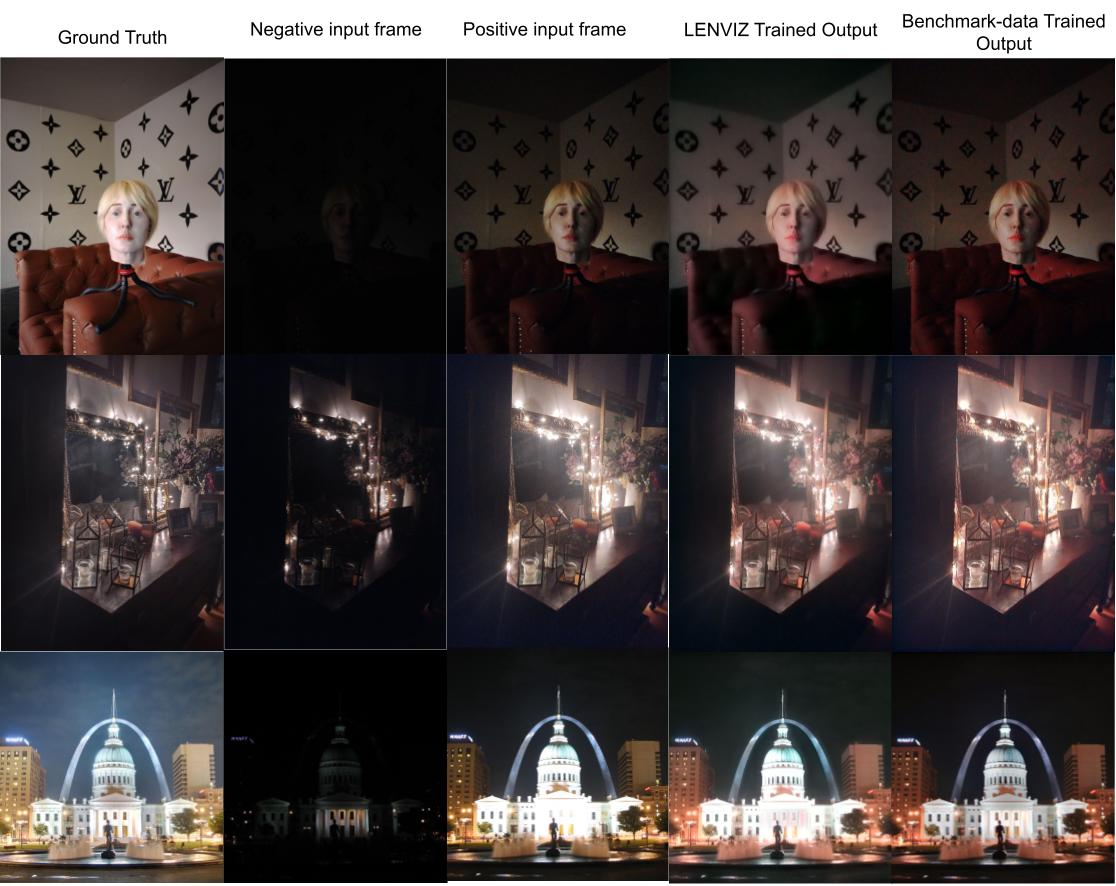}
   \vspace{-2em} \caption{Example of failed cases where users preferred Benchmark-trained model output (MobileMEF) compared to LENVIZ trained model output supported with GT and input frames (dark and bright).}  \vspace{-1em} 
   \label{fig:supp_failed_case}
\end{figure*}

We also include a dedicated analysis of failure cases to provide a more comprehensive understanding of LENVIZ data trained model's limitations and to guide future research. Fig~\ref{fig:supp_failed_case} showcases these instances with a side-by-side comparison of our data trained model output against the benchmark data trained model output. In these specific cases, the user study revealed a preference for the benchmark trained model output due to its superior color, contrast, detail, and lower noise. Upon analysis, we observed that while the outputs of our data trained model output and the benchmark are visually quiet similar, the quality difference is largely attributable to our model's early stopping. Due to time constraints, the model was trained only 100 epoch using our dataset and with sheer the size of the dataset the training time was much longer not allowing us to wait for full convergence. Based on our prior experience with similar models, we are confident that continued training would enable our data trained model to produce results that not only match but also surpass the benchmark trained model output quality.

\subsubsection{Model Generalizability}\label{generalizability}
\begin{figure*}[!h]
   \centering
   \includegraphics[width=1\linewidth]{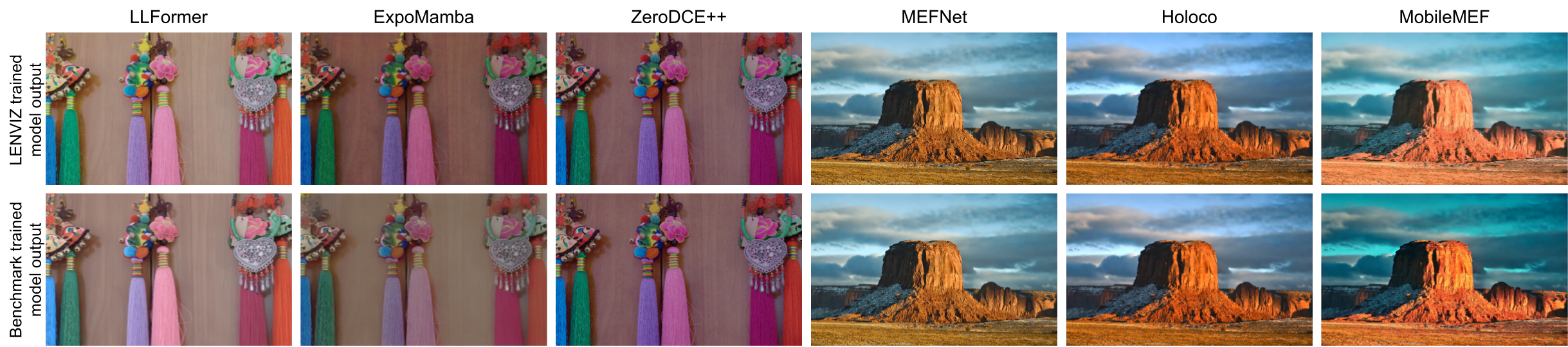}
   \vspace{-2em} \caption{Output samples from models trained with LENVIZ data vs Benchmark-data (LOL/SICE) using the benchmark test data. Model's trained on LENVIZ demonstrate comparable or superior image quality even when tested on data captured by different (unseen) cameras, showcasing the model's generalizability after the model is trained using LENVIZ dataset}  \vspace{-1em} 
   \label{fig:generalizability}
\end{figure*}

The examples in Fig~\ref{fig:generalizability} showcase the improved performance of the SOTA models when trained on LENVIZ data vs benchmark (LOL/SICE) dataset. For fairness of comparison, the test data used in this study was benchmark (LOL/SICE). We can observe that the model output when trained on LENVIZ data demonstrates improved brightness, contrast, texture, and sharpness. This provides string empirical evidence that our dataset, being captured with 3 different camera module under fixed camera settings, enables models to learn highly robust and generalizable features. The unprecedented scale and diversity of our dataset are key factors in its effectiveness as a training tool for low-light enhancement models across a variety of camera hardware and scene types.



\end{document}